\documentclass[10pt,journal]{IEEEtran}
\usepackage{enumitem,calc}
\usepackage{graphicx}
\usepackage{subfigure}
\usepackage{amsmath}
\usepackage{array}
\usepackage{subfig}
\usepackage{tabu}
\usepackage{amssymb}
\usepackage{multirow}
\usepackage{booktabs}
\usepackage{cite}
\usepackage{xcolor}
\usepackage{soul}
\usepackage{threeparttable}
\usepackage{rotating}
\soulregister\cite7
\soulregister\ref7

\ifCLASSINFOpdf
\else
\fi
\begin{document}
	
	\title{MINet: Multi-scale Interactive Network for Real-time Salient Object Detection of Strip Steel Surface Defects}
	
	\author{Kunye~Shen,
		Xiaofei~Zhou,
		and Zhi~Liu,~\IEEEmembership{Senior Member,~IEEE}
		\thanks{\textit{Corresponding authors: Xiaofei Zhou and Zhi Liu.}}	
		\thanks{Kunye Shen and Zhi Liu are with the Key Laboratory of Specialty Fiber Optics and Optical Access Networks, Joint International Research Laboratory of Specialty Fiber Optics and Advanced Communication, Shanghai Institute for Advanced Communication and Data Science, School of Communication and Information Engineering, Shanghai University, Shanghai 200444, China (email: KunyeShen@outlook.com; liuzhisjtu@163.com).}
		\thanks{Xiaofei Zhou is with the School of Automation, Hangzhou Dianzi University, Hangzhou 310018, China (e-mail: zxforchid@outlook.com).}}
	
	\maketitle
	
	\begin{abstract}
	The automated surface defect detection is a fundamental task in industrial production, and the existing 
	saliency-based works overcome the challenging scenes and give promising detection results. However, the cutting-edge efforts often suffer from large parameter size, heavy computational cost, and slow inference speed, which heavily limits the practical applications. 
	To this end, we devise a multi-scale interactive (MI) module, which employs depthwise convolution (DWConv) and pointwise convolution (PWConv) to independently extract and interactively fuse features of different scales, respectively. Particularly, the MI module can provide satisfactory characterization for defect regions with fewer parameters. Embarking on this module, we propose a lightweight Multi-scale Interactive Network (MINet) to conduct real-time salient object detection of strip steel surface defects. 
	Comprehensive experimental results on SD-Saliency-900 dataset, which contains three kinds of strip steel surface defect detection images (\emph{i.e.}, inclusion, patches, and scratches), demonstrate that the proposed MINet presents comparable detection accuracy with the state-of-the-art methods while running at a GPU speed of 721FPS and a CPU speed of 6.3FPS for 368$\times$368 images with only 0.28M parameters. The code is available at https://github.com/Kunye-Shen/MINet.
	\end{abstract}
	

	\IEEEpeerreviewmaketitle
	
	\section{Introduction}
	\label{sec:introduction}
	\IEEEPARstart{S}{teel} is one of the fundamental materials in industrial production, where the defects of steel surface such as inclusion, patches, and scratches have a severe impact on the aesthetics and reliability of industrial products. Therefore, strip steel surface defect detection is an indispensable part of steel industrial production. Meanwhile, due to the complex environment of strip steel production, defect detection accuracy is highly susceptible. In addition, the detection accuracy, model size and inference speed collectively determine the practical application capability of detection methods.
	
	In recent years, with the rapid development of deep learning technologies, especially the convolutional neural networks (CNNs), defect detection achieves great progress. Particularly, many saliency-based defect detection models \cite{edrnet, dacnet, eminet, wan2023lfrnet, wan2023canet} present promising detection results, where the traditional cumbersome backbones including VGG\cite{vgg} and ResNet\cite{resnet} provide effective multi-scale features. Considering the challenges of defect detection including low contrast between defects and backgrounds, various shapes and scales of defect, and low illumination, the multi-scale strategy is widely utilized by many efforts \cite{dong2019pga, dacnet, eminet}, which provides rich contextual information for defects regions. 
	However, they often suffer from large computational overhead and slow running speed, which inevitably overtaxes the limited storage and computational capabilities of industrial devices. 
	Therefore, it would be meaningful work that design a saliency-based lightweight network with satisfactory detection accuracy for strip steel surface defect detection.


	\begin{figure}[!t]
		\centerline{\includegraphics[width=\columnwidth]{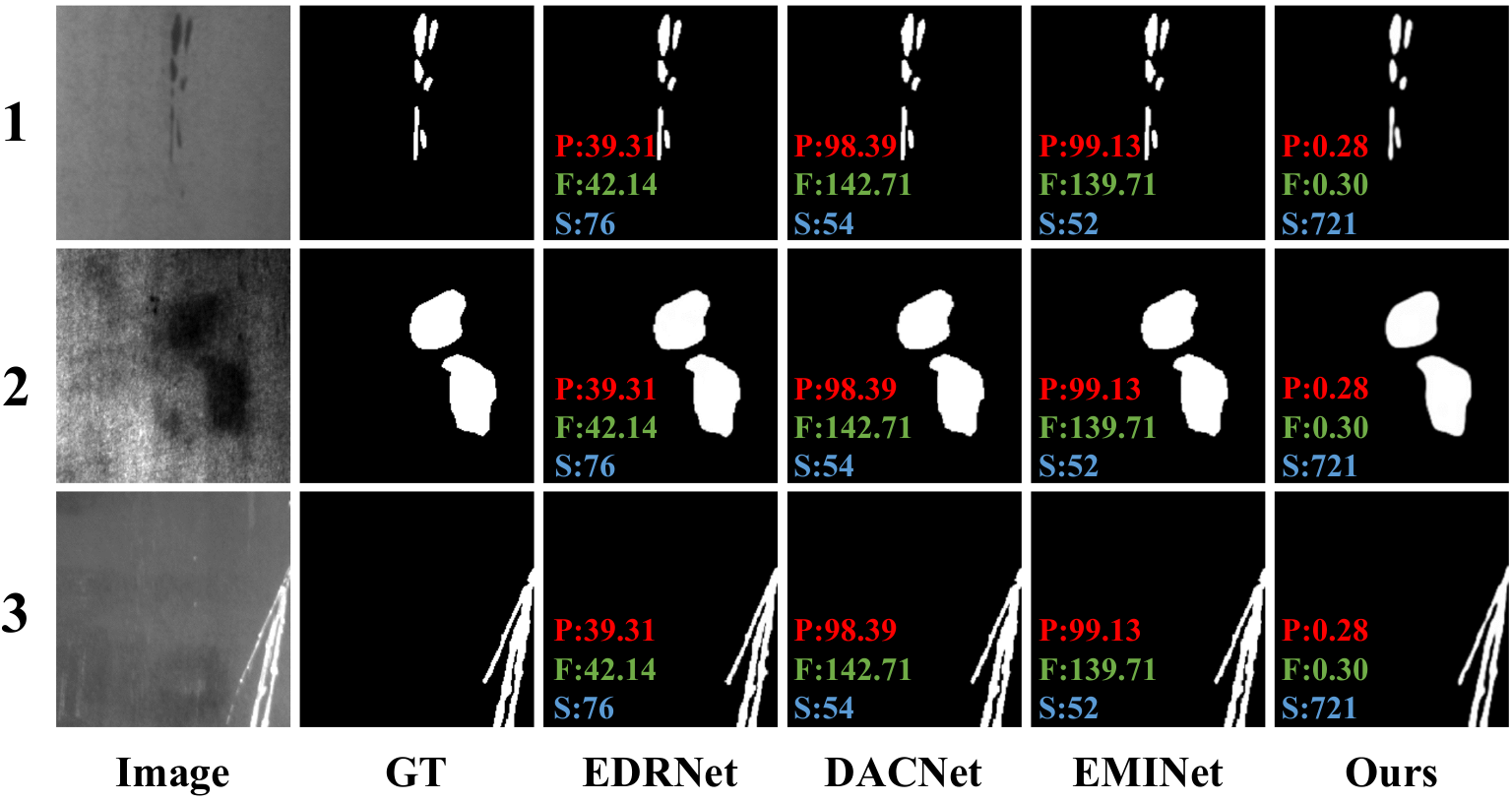}}
		\caption{Visual comparison between our MINet and 3 state-of-the-art strip steel surface defect detection methods. Parameters(M), FLOPs(G), and Speed(FPS) are marked in \textcolor{red}{red}, \textcolor{green}{green}, and \textcolor{blue}{blue}, respectively.}
		\label{Qul_Cmp_1}
	\end{figure}
	
	Fortunately, there are many researchers have devoted their efforts to the design of lightweight networks such as MobileNet\cite{mobilenet} and ShuffleNets\cite{zhang2018shufflenet}. 
	Meanwhile, we notice that the depthwise separable convolution (DSConv) \cite{mobilenet} has attracted more and more concerns in many lightweight models\cite{SAMNet,EDN}. But we can also find that, to acquire the multi-scale information of objects, the existing lightweight efforts often adopt a series of DSConv layers, where they employ element-wise summation to fuse features of different scales. In this way, the features may be weakened. Besides, concatenation will not diminish the characterization ability of multi-scale features, where each scale feature retains its original information. However, directly introducing concatenation operation after DSConv will increase computational costs, which contradicts the goal of real-time tasks. In addition, a single DSConv cannot provide rich contextual information for the strip steel surface defects. Therefore, it is an interesting idea that embedding the multi-scale strategy into the DSConv, which not only provides an effective representation for defect regions but also is endowed with lower computational overhead.
	


	Motivated by the aforementioned descriptions, we first design a Multi-scale Interactive (MI) module to expand the traditional DSConv, as shown in Fig.~\ref{MI module}. The key advantage of MI module is that we embed the multi-scale strategy into the DSConv, where depthwise convolution (DWConv) and pointwise convolution (PWConv) are used to extract and aggregate multi-scale features, respectively. Meanwhile, we notice that the prior arts\cite{HVPNet,SAMNet,FSMINet} successfully employed convolutions with different dilation rates to enhance the characterization of objects. Therefore, the MI module first employs four DWConv blocks with different dilation rates to extract multi-scale features, which are with different receptive fields. Then, to promote the interaction of the multi-scale features of DWConv, we attempt to combine these features in a hierarchical way. Concretely, we take the feature of each channel from the four-scale features separately, and concatenate the features of the same channel index, yielding a series of features with four channels. After that, these four-channel features are initially fused by PWConv blocks, and we can acquire a set of single-channel features, which are concatenated together and further processed by PWConv block. In this way, we can obtain the fused feature,  
	which can provide an effective depiction of defects. Finally, to preserve the initial information, the fused feature combines the original input in a summation way, yielding the enhanced input feature. 
	Moreover, we provide the analysis of computational cost in Sec.~\ref{Computational cost analysis}. It can be found that the computational cost of our MI module is lower than that of directly adopting concatenation after DSConv.
	Following this way, we can not only significantly reduce the computational cost but also harvest the effective features which can give an powerful representation for defects.
	
	Secondly, embarking on the MI module, we develop a real-time backbone, as shown in Fig.~\ref{backbone}. Concretely, following the architecture of ResNet-50\cite{resnet}, we construct the MI-based real-time backbone containing five encoder blocks, where the first encoder block is a conventional 3$\times$3 convolution layer and the last four encoder blocks are built on the MI module. Particularly, to achieve low latency\cite{vasu2023mobileone}, we arrange the MI modules in a linear way within each encoder block. With this backbone, we can acquire effective multi-scale contextual features. Lastly, we propose a lightweight saliency model, namely Multi-scale Interactive Network (MINet), to conduct real-time strip steel surface defect detection, which achieves a trade-off between detection performance and computational efficiency. Specifically, the MINet presented in Fig.~\ref{MINet} adopts an encoder-decoder architecture, where the encoder adopts the MI-based real-time backbone and each decoder block consists of two DSConv layers with different dilation rates. 
The MINet achieves comparable detection accuracy with the state-of-the-art models while running at 721FPS on an NVIDIA GTX 2080Ti GPU and 6.3FPS on a i9-9900X CPU for 384$\times$384 images. In addition, MINet only contains 0.28M parameters and 0.30G FLOPs. 




In summary, our contributions are as follows:\begin{enumerate}
\item[$\bullet$]
We propose a novel Multi-scale Interactive (MI) module, which embeds the multi-scale strategy into the DSConv and has a low computational cost. Particularly, the MI module employs the DWConv and PWConv to extract and aggregate the multi-scale features, respectively, which can provide an effective depiction of defects. 


\item[$\bullet$]
We propose a MI-based real-time backbone containing five encoder blocks, where each of the last four encoder blocks arranges the MI modules in a linear way to acquire low latency. Embarking on the backbone, we construct a lightweight Multi-scale Interactive Network (MINet) for real-time salient object detection of strip steel surface defects, which achieves a good balance between detection accuracy and computational efficiency. 




\item[$\bullet$]
We conduct comprehensive experiments on the challenging dataset, and the experimental results firmly prove MINet's comparable detection accuracy with fewer parameters and higher running speed when compared with the state-of-the-art models.

\end{enumerate}

The remaining of this article are summarized as follows: In Sec.~\ref{Related Work}, we introduce the related works of strip steel surface defect detection and real-time neural networks. After that, we detail the components of our MINet in Sec.~\ref{Methodology}, and show the comparison results in Sec.~\ref{Experiments}. Finally, we conclude the entire paper in Sec.~\ref{Conclusion}.

\section{Related Work}\label{Related Work}
\subsection{Surface Defect Detection}
In recent years, there has been a growing interest in automated surface defect detection, and lots of detection methods have been designed. In the following, we will give a brief overview of the existing surface defect detection methods. Specifically, to localize the defect regions, some methods attempt to provide bounding boxes to cover the defect regions. For example, in \cite{ni2021attention}, attention neural network with variable receptive fields is introduced to flexibly suppress background noise and improve the defect detection accuracy in rail surface. Besides, Song \emph{et al.} \cite{song2023steel} promoted detection accuracy through the improved Faster R-CNN method, which incorporates deformable convolution and Region-of-Interest (RoI) alignment. In \cite{yeung2022efficient}, an adaptively balanced feature fusion method is designed to combine features with appropriate weights, thereby enhancing the characterization ability of the feature maps for defects. In the above three studies, researchers regarded defect detection as a key-point estimation task. Furthermore, some methods try to conduct pixel-wise detection of surface defects. For example, in \cite{dong2019pga}, Song \emph{et al.} proposed a pyramid feature fusion and global context attention network to fuse multi-level features into different resolution features and promote the information propagation among different resolution features. In \cite{ma2023shape}, a one-shot unsupervised domain adaptation framework is proposed to promote the method's generalization ability. Furthermore, to address the challenge of limited samples in the target domain, they introduced a novel framework that requires only a single sample from the target domain to enhance the robustness of the algorithm. In addition, Liu \emph{et al.} \cite{liu2022tas} introduced a multi-level feature extraction module to capture more contextual information in steel surface environment. In this way, the proposed network can better identify small defect regions.

In addition, some methods regard the defect detection as the salient object detection task. For example, in \cite{edrnet}, an end-to-end encoder-decoder residual network (EDRNet) is proposed to detect the strip steel surface defects. In \cite{dacnet}, Zhou \emph{et al.} proposed a dense attention-guided cascaded network (DACNet) by deploying multi-resolution convolutional branches. In \cite{eminet}, Zhou \emph{et al.} further proposed an edge-aware multi-level interactive network, which relies on the interactive feature integration and the edge-guided saliency fusion. Besides, Han \emph{et al.} \cite{han2022two} designed a two-stage edge reuse network, which executes prediction and refinement successively. Ding \emph{et al.} \cite{ding2022cross} proposed a cross-scale edge purification network, where the cross-scale calibration module and cross-scale feature interweaving module are deployed to explore the correlations of different-scale features. Following the aforementioned research, we aim to conduct relevant studies on real-time salient object detection of surface defect detection.




\subsection{Real-time Neural Network}

In recent years, to promote the application of neural networks in real scenarios, real-time neural networks attract more and more concerns, which plays an important role in a scenario with limited computational resources. There are many efforts have been devoted to the design of real-time neural networks. For example, in \cite{squeezenet}, Iandola \emph{et al.} proposed a small CNN architecture, namely SqueezeNet, which provides AlexNet-level performance on ImageNet and has $50\times$ fewer parameters than AlexNet. In \cite{mobilenet}, Howard \emph{et al.} proposed an efficient streamlined model MobileNets, which employs the depthwise separable convolution (DSConv) to construct lightweight neural networks. In \cite{efficientnet}, Tan \emph{et al.} proposed an effective compound scaling method to scale up a baseline network to any target resource constraints, which also maintains the model efficiency. Besides, Yu \emph{et al.}\cite{BiSeNet} introduced a novel bilateral segmentation network to achieve a balance between segmentation performance and inference speed. In\cite{LEDNet}, Wang \emph{et al.} employed the channel split and shuffle to equip the encoder, and they designed an attention pyramid network as the decoder. Wu \emph{et al.}\cite{CGNet} proposed a context guided block to capture contextual information at all stages of the network. Meanwhile, we can find that the DSConv \cite{mobilenet} has been successfully used by many efforts \cite{SAMNet, HVPNet} in recent years, which largely elevates the efficiency of the neural networks. Therefore, in this paper, we build our MI module by using the DSConv.


\section{Methodology}\label{Methodology}
In this section, we first review the composition and computational cost of the DSConv. After that, in Sec.~\ref{Preliminary}, we briefly present three multi-scale feature fusion methods and point out the problem brought by summation or concatenation within fusion process. Then, in Sec.~\ref{Multi-scale interactive module}, we describe the details of the MI module. Next, we detail the MI-based real-time backbone and build the MINet. Lastly, in Sec.~\ref{Loss function}, we present the loss function.

\subsection{Preliminary}\label{Preliminary}
As mentioned in \cite{mobilenet}, DSConv consists of depthwise convolution (DWConv) and pointwise convolution (PWConv). Specifically, for the DWConv operation, the input image $\mathbf{I}\in \mathbb{R}^{c\times h\times w}$ is processed by a $k\times k$ convolutional layer $\mathbf{W}_{D}\in \mathbb{R}^{c\times k\times k}$ to obtain the feature $\mathbf{F}\in \mathbb{R}^{c\times h\times w}$. Successively, the DSConv conducts PWConv to remedy the lack of inter-channel interaction of the DWConv, where a $1\times1$ convolutional layer $\mathbf{W}_{P}\in \mathbb{R}^{c\times 1\times 1}$ is used to get the final output of DSConv, namely $\mathbf{O}\in \mathbb{R}^{c\times h\times w}$.

The compuatational cost of DSConv can be calculated as:
\begin{equation}\label{eq1}
\begin{array}{lc}
k^{2}\cdot c\cdot w\cdot h + c^{2}\cdot w\cdot h
\end{array},
\end{equation}
where $w$ and $h$ represent the width and high of feature map, respectively. $c$ denotes the channel numbers. 

\begin{figure}[h] \centerline{\includegraphics[width=\columnwidth]{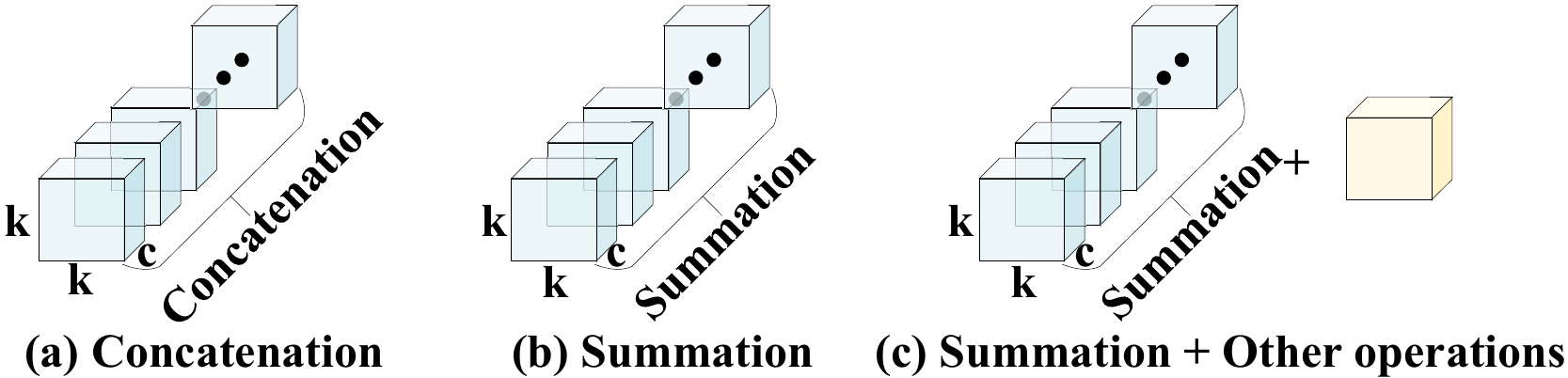}}
\caption{Illustration of three types of the fusion of multi-scale features, including (a) Concatenation, (b) Summation, and (c) Summation with other operations.}
\label{Fea_Com}
\end{figure}

\begin{figure*}[h]
\centerline{\includegraphics[width=\textwidth]{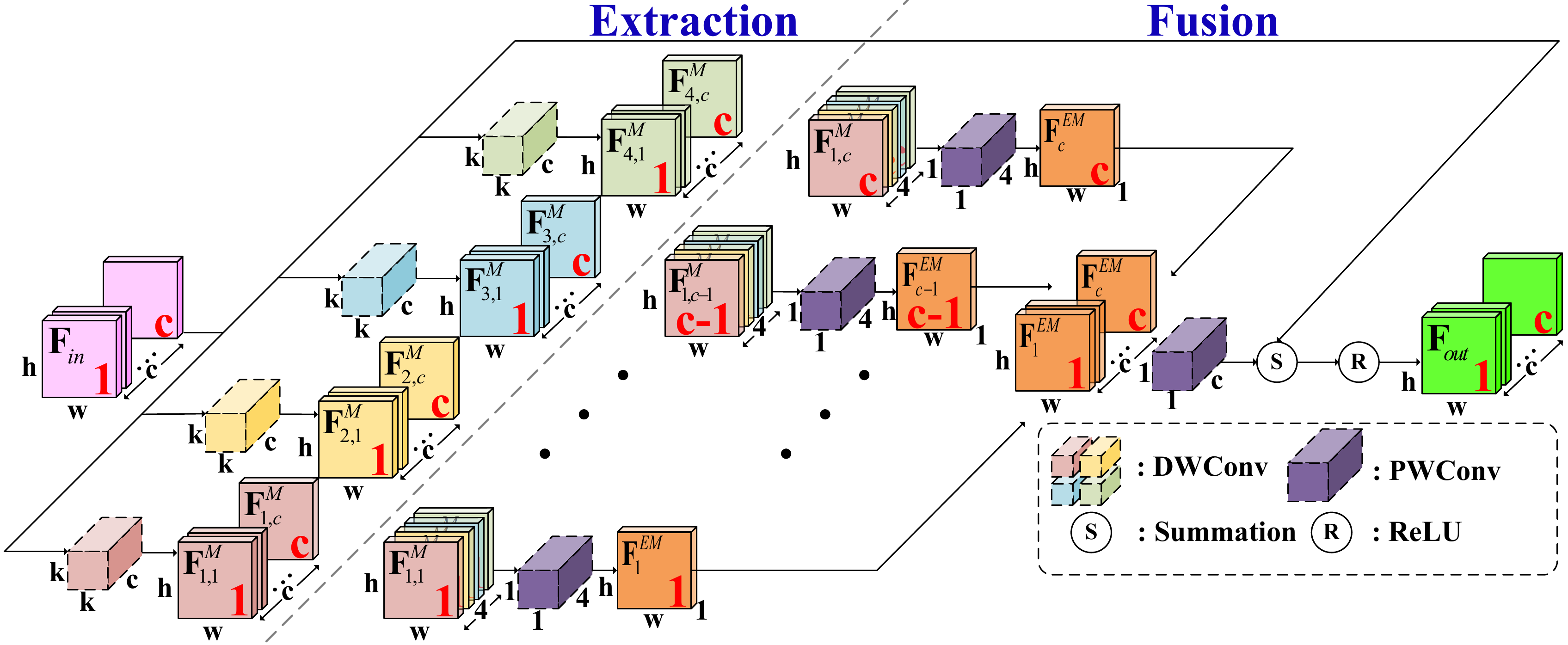}}
\caption{Illustration of the proposed MI module. The input feature $\mathbf{F}_{in}\in \mathbb{R}^{c\times h\times w}$ is processed by four DWConvs ($f_{DW}\in \mathbb{R}^{c\times 3\times 3}$) to acquire multi-scale features $\{\mathbf{F}_{i}^{M}\}_{i=1}^{4}\in \mathbb{R}^{c\times h\times w}$ during the first stage. After that, enhanced multi-scale features $\{\mathbf{F}_{i}^{EM}\}_{i=1}^{c}\in \mathbb{R}^{1\times h\times w}$ are obtained by PWConv ($f_{PW}\in \mathbb{R}^{4\times 1\times 1}$). And then, PWConv ($f_{PW}\in \mathbb{R}^{c\times 1\times 1}$) is used to fuse enhanced multi-scale features across channels. Following the residual structure, we finally obtain the $\mathbf{F}_{out}\in \mathbb{R}^{c\times h\times w}$ of our MI module.}
\label{MI module}
\end{figure*}

We notice that to combine multi-scale features, many efforts have been designed. In Fig.~\ref{Fea_Com}(a), it is the concatenation operation, where it fully preserves the original information. However, the output of concatenation operation will cause a huge amount of computation, and thus, the concatenation operation cannot be directly used in real-time tasks without any modifications. Besides, some real-time works tend to combine features by using cost-effective methods (\emph{e.g.}, summation), as shown in Fig.~\ref{Fea_Com}(b). Nonetheless, as mentioned in \cite{SAMNet}, if features from different receptive fields are summarized directly, the resulting information may be diluted. Moreover, the work\cite{SAMNet} alleviates this issue by introducing other operations (\emph{i.e.,} channel-wise attention mechanism, spatial attention mechanism, and volumetric attention), as presented in Fig.~\ref{Fea_Com}(c), which achieved satisfactory effects, but it also increases the computational cost. To address this problem, we proposed our multi-scale interactive (MI) module, as shown in Fig.~\ref{MI module}.

\subsection{Multi-scale interactive module}\label{Multi-scale interactive module}
\subsubsection{Architecture}
According to the effort \cite{mobilenet}, we can find that the DSConv factorizes a standard convolution into a DWConv and a PWConv, where the former acts as a filter to extract features and the latter linearly combines the outputs of the former. Therefore, embarking on the DSConv, we build our multi-scale interactive (MI) module, which attempts to embed the multi-scale strategy into the DSConv. Here, the MI module utilizes the DWConv to perform ``Feature Extraction'' and employs the PWConv to conduct ``Feature Fusion''. 
Besides, the dilated convolution-based multi-scale strategy has been successfully employed by \cite{HVPNet,SAMNet,FSMINet}, which endows features with different receptive fields. This can effectively promote the characterization of defects. Therefore, the feature extraction part first deploys four DWConvs with different dilation rates (\emph{i.e.,} 1, 2, 4, and 8), and obtains a set of multi-scale features $\{\mathbf{F}_{i}^{M}\}_{i=1}^{4}\in \mathbb{R}^{c\times h\times w}$. We should note that the multi-scale features refer to the features with four different receptive fields, and their spatial resolution is same. The process can be defined as follows,
\begin{equation}\label{eq2}
\begin{array}{lc}
\mathbf{F}_{i}^{M} = f_{DW}^{2^{i-1}}(\mathbf{F}_{in}) & i=1,2,3,4 \\
\end{array},
\end{equation}
where $f_{DW}^{2^{i-1}}$ is a $3\times3$ DWConv with a dilation rate of $2^{i-1}$. $\mathbf{F}_{in} \in \mathbb{R}^{c\times h\times w}$ denotes the input of our MI module. Here, due to the limitation of DWConv, namely the lack of inter-channel interactions, the acquired features cannot sufficiently explore the channel correlation. Therefore, we introduce the PWConv into the subsequent operation, namely ``Feature Fusion''. 

Formally, we employ the PWConv to fuse the outputs of the ``Feature Extraction''. Here, as shown in Fig.~\ref{MI module}, we take a stepwise fusion approach. Firstly, we split each of the multi-scale features $\mathbf{F}_i^M\in \mathbb{R}^{c\times h\times w}$ along the channel dimension to obtain $\{\mathbf{F}_{i,j}^{M}\}_{j=1}^c$, where $\mathbf{F}_{i,j}^{M}\in \mathbb{R}^{1\times h\times w}$. In this way, we can obtain four-scale feature sets $\{\{\mathbf{F}_{1,j}^M\}_{j=1}^{c},\{\mathbf{F}_{2,j}^M\}_{j=1}^{c},\{\mathbf{F}_{3,j}^M\}_{j=1}^{c},\{\mathbf{F}_{4,j}^M\}_{j=1}^{c}\}$. 
After that, we integrate the feature maps of various receptive fields, which are with the same channel index, yielding enhanced multi-scale features $\{\mathbf{F}_{j}^{EM}\}_{j=1}^{c}$. This process can be formulated as
\begin{equation}\label{eq3}
\begin{array}{lc}
\mathbf{F}_{j}^{EM} = f_{PW}(Cat(\mathbf{F}_{1,j}^{M}, \mathbf{F}_{2,j}^{M}, \mathbf{F}_{3,j}^{M}, \mathbf{F}_{4,j}^{M})) & j=1,2, \cdots, c\\
\end{array},
\end{equation}
where $f_{PW}$ means the $1\times1$ PWConv (channel number = 4), and $Cat$ denotes the concatenation operation. Following this way, we can preliminarily combine the multi-scale features. Particularly, with the concatenation operation, multi-scale features can avoid the dilution issue caused by the summation operation. 

After that, we concatenate the enhanced multi-scale features $\{\mathbf{F}_{j}^{EM}\}_{j=1}^c$, which will be further processed by PWConv. In this way, we can emphasize the feature interactions among different channels and different scales simultaneously. 
Besides, we adopt the residual structure to combine the original input $\mathbf{F}_{in}$, yielding the final output of our MI module, namely $\mathbf{F}_{out}$. The entire process can be written as 
\begin{equation}\label{eq4}
\begin{array}{lc}
\mathbf{F}_{out} = f_{R}(f_{PW}(Cat(\mathbf{F}_{1}^{EM}, \cdots, \mathbf{F}_{c}^{EM}))+\mathbf{F}_{in})\\
\end{array},
\end{equation}
where $f_{PW}$ means the $1\times1$ PWConv (channel number = $c$), and $\mathbf{F}_{out}\in \mathbb{R}^{c\times h\times w}$ represents the output of our MI module. 
$f_{R}$ and $+$ mean ReLU function and pixel-wise summation, respectively.

\subsubsection{Computational cost analysis}\label{Computational cost analysis}



According to Fig.~\ref{MI module} and the aforementioned descriptions, we can give the computational cost of MI module, namely
\begin{equation}\label{eq5}
\begin{array}{lc}
4\cdot k^{2}\cdot c\cdot w\cdot h + c\cdot 4\cdot w\cdot h + c^{2}\cdot w\cdot h
\end{array}.
\end{equation}

Meanwhile, to demonstrate the efficiency of our MI module, we make some modifications for our MI module, namely a variant of MI module. Concretely, we first deploy four DSConvs with four different dilation rates to extract features, and then combine them via concatenation or element-wise summation, as presented in Fig.~\ref{Fea_Com} (a) and (b). In this way, we can not only obtain four features with different receptive fields but also acquire the enhanced multi-scale feature. According to Eq.~\ref{eq1}, the corresponding computational cost of this design can be written as follows
\begin{equation}\label{eq6}
\begin{array}{lc}
4\cdot(k^{2}\cdot c\cdot w\cdot h + c^{2}\cdot w\cdot h)
\end{array}.
\end{equation}

By comparing Eq.~\ref{eq5} and Eq.~\ref{eq6}, we can find Eq.~\ref{eq5} is less than Eq.~\ref{eq6} when c is greater than $1$. 
Therefore, we can theoretically confirm that our MI module has a more efficient lightweight structure when compared with the existing efforts targeting multi-scale feature extraction. 
The ablation results will be illustrated in Sec.~\ref{Ablation analysis}.

\subsection{Network Architecture}\label{Network Architecture}
The existing surface defect detection efforts\cite{edrnet,dacnet,eminet} are built on the ResNet backbone\cite{resnet}. However, the vast number of parameters and floating-point computations in ResNet significantly increase the processing time of the proposed network. Therefore,  we attempt to build a real-time backbone by utilizing our MI module, as presented in Sec.~\ref{MI-based real-time backbone}. Embarking on this, we deploy an encoder-decoder network for defect detection, as described in Sec.~\ref{MINet architecture}.

\begin{figure*}[ht] \centerline{\includegraphics[width=\textwidth]{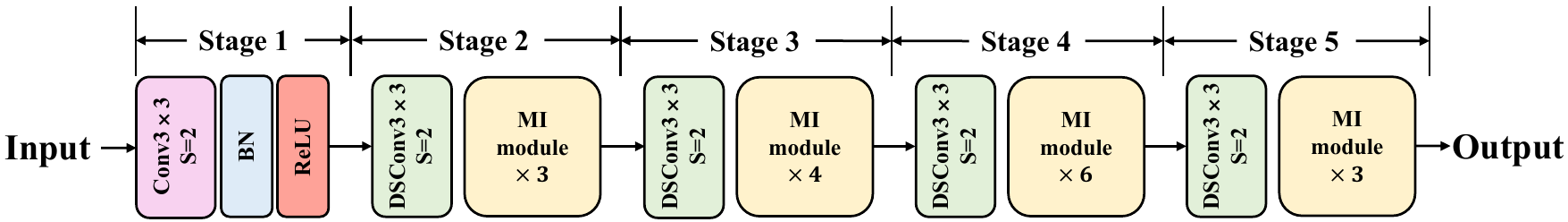}}
\caption{Architecture of the MI-based real-time backbone. Our MI-based real-time backbone comprises five stages.}
\label{backbone}
\end{figure*}

\subsubsection{MI-based real-time backbone}\label{MI-based real-time backbone}
Embarking on our MI module, we have further developed a real-time backbone for the strip steel surface defect detection. In terms of architecture, our MI-based real-time backbone can be easily divided into five distinct parts. The first part comprises convolution, batch normalization layer, and ReLU function. The subsequent parts consist of a DSConv (stride=2) and several of our MI modules. Next, we will discuss the details and functions of each part of our MI-based real-time backbone.

Following previous works\cite{resnet,vgg}, we utilize five encoder stages (or encoder blocks) to capture contextual information, as illustrated in Fig.~\ref{backbone}. Specifically, the first stage consists of a convolutional layer (stride=2), a batch normalization (BN) layer, and a ReLU activation function. Here, different from the large convolutional kernel (\emph{i.e.}, $7\times7$) of the $1^{st}$ encoder block in ResNet\cite{resnet}, we set the convolution kernel size of stage1 to $3\times3$, which can effectively reduce the computational cost. The following four stages all consist of a $3\times3$ DSConv (stride=2) and varying numbers of MI modules, where the former is used to downsample the spatial resolution and expand the channel dimension of features and the latter is utilized to extract features.  

Besides, inspired by ResNet50\cite{resnet}, we set the number of MI module units in stage2, 3, 4, and 5 to 3, 4, 6, and 3, respectively. Here, we should note that in our real-time backbone, MI modules are linearly arranged within the same stage, which is beneficial for acquiring low latency\cite{vasu2023mobileone}. Besides, the multi-scale strategy in each MI module promotes the acquisition and integration of multi-scale features, and thus, our MI-based backbone can give a good depiction of defects. Detail information such as resolution of each stage is illustrated in Table~\ref{backbone setting}. Following the feature extraction process, namely the MI-based real-time backbone, we can successfully obtain contextual information $\{\mathbf{F}_{i}\}_{i=1}^{5}$ for subsequent decoding process.

\begin{table}[hb]
\centering
\caption{Details of MI module-based Real-time Backbone}
\footnotesize
\renewcommand{\arraystretch}{1.0}
\renewcommand{\tabcolsep}{4.5mm}
\begin{tabular}{c|c|c|c|c}
\toprule
\textbf{Stage} & \textbf{Resolution} & \textbf{Module} & \textbf{\#M} & \textbf{\#F} \\
\midrule
\multirow{2}[2]{*}{1} & 368$\times$368 & \multirow{2}[2]{*}{Conv3$\times$3} & \multirow{2}[2]{*}{1} & \multirow{2}[2]{*}{16} \\
& 184$\times$184 &       &       &  \\
\midrule
\multirow{2}[2]{*}{2} & 184$\times$184 & DSConv3$\times$3 & 1     & 32 \\
& 92$\times$92 & MI module & 3     & 32 \\
\midrule
\multirow{2}[2]{*}{3} & 92$\times$92 & DSConv3$\times$3 & 1     & 64 \\
& 46$\times$46 & MI module & 4     & 64 \\
\midrule
\multirow{2}[2]{*}{4} & 46$\times$46 & DSConv3$\times$3 & 1     & 96 \\
& 23$\times$23 & MI module & 6     & 96 \\
\midrule
\multirow{2}[2]{*}{5} & 23$\times$23 & DSConv3$\times$3 & 1     & 128 \\
& 12$\times$12 & MI module & 3     & 128 \\
\bottomrule
\end{tabular}%
\begin{tablenotes}
\footnotesize
\item ``\#M`` means the number of modules, and the ``\#F`` means the number of the convolution filters (\emph{i.e.,} channels).
\end{tablenotes}
\label{backbone setting}%
\end{table}%

\begin{figure}[htb!]
\centerline{\includegraphics[width=\columnwidth]{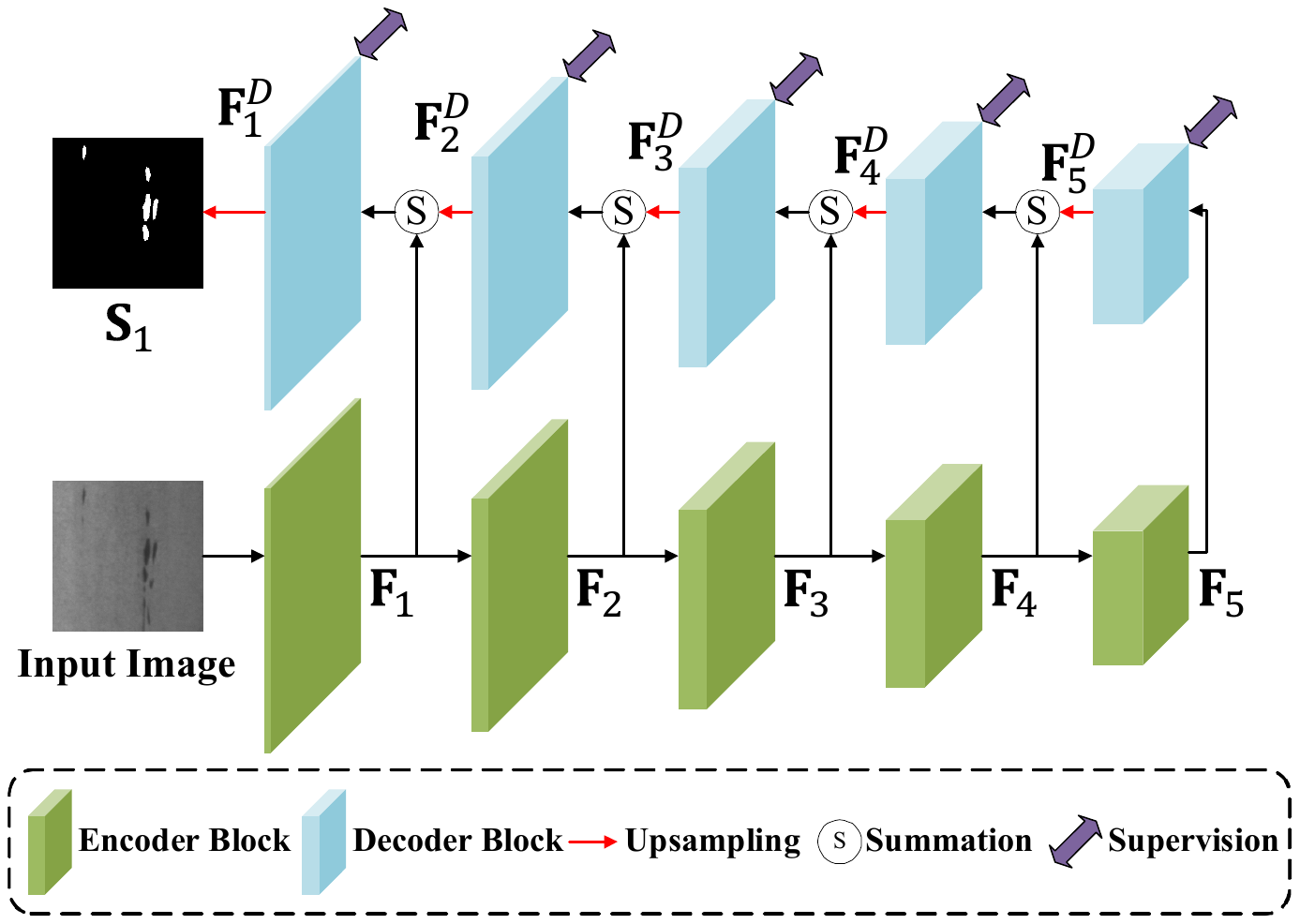}}
\caption{Overall architecture of our MINet. $\{\mathbf{F}_i\}_{i=1}^5$ and $\{\mathbf{F}_i^D\}_{i=1}^5$ represents the outputs of Encoder-$i$ and Decoder-$i$, respectively.}
\label{MINet}
\end{figure}
\subsubsection{MINet architecture}\label{MINet architecture}
Drawing on previous works\cite{BASNet,U2Net,PoolNet}, we adopt the classical encoder-decoder architecture to build our real-time model. Concretely, we employ our MI-based real-time backbone as the encoder to extract multi-level contextual information, namely multi-scale features $\{\mathbf{F}_i\}_{i=1}^5$, as illustrated in Fig.~\ref{MINet}. 
Subsequently, we deploy the decoder consisting of five decoder blocks to give the final defect detection results. Concretely, each decoder block contains a $3\times3$ DSConv (dilation rate=2, stride=1) and a $3\times3$ DSConv (dilation rate=1, stride=1). This entire decoding process can be formulated as follows
\begin{equation}\label{eq7}
\mathbf{F}_{i}^{D} = \left \{\begin{array}{lc}
f_{D}^i(\mathbf{F}_{i}+Up_{\times2}(\mathbf{F}_{i+1}^{D})) & i=1,2,3,4\\
f_{D}^i(\mathbf{F}_{i}) & i=5 \\
\end{array},
\right.
\end{equation}
where $\{\mathbf{F}_{i}\}_{i=1}^{5}$ and $\{\mathbf{F}_{i}^{D}\}_{i=1}^{5}$ represent the $i^{th}$ encoder and decoder features of the encoder-$i$ and decoder-$i$, respectively. $+$ is element-wise summation, and $Up_{\times2}$ represents $2\times$ upsampling operation, which is actually the bilinear interpolation. $f_{D}^i$ denotes the $i^{th}$ decoder block. 

Then, we deploy the output layer on each decoder feature to generate a prediction map, where the output layer consists of a dropout layer (rate=0.1), a $1\times1$ convolution layer (stride=1, channel number=1), and a sigmoid activation function. Here, to match the input spatial resolution, the bilinear interpolation based upsampling operation is utilized to resize the spatial resolution of the detection result, which can be formulated as follows
\begin{equation}\label{eq8}
\begin{array}{lc}
\mathbf{S}_{i} = Up_{\times2^{i}}(f_{O}(\mathbf{F}_{i}^{D})) & i=1,2,3,4,5\\
\end{array},
\end{equation}
where $\{\mathbf{S}_{i}\}_{i=1}^{5}$ are detection results with different resolutions, and $f_{O}$ is the output layer. $Up_{\times2^{i}}$ means $2^{i}\times$ upsampling operation. Here, $\mathbf{S}_{1}$ boasts the highest resolution, and thus we choose it as the final defect detection result.

\subsection{Loss function}\label{Loss function}
Deep supervision has been widely employed by previous works\cite{edrnet,dacnet,eminet}, where it can comprehensively evaluate the prediction results. Therefore, our work also adopts the deep supervision strategy. 
Here, we adopt the BCE loss and SSIM loss\cite{ssim} to compute the loss function. It should be noted that SSIM loss steers the network to pay more attention to the boundaries of defect regions, which is beneficial for increasing detection accuracy. 
The hybrid loss function can be expressed as follows:
\begin{equation}\label{eq9}
\begin{array}{lc}
\mathcal{L}_{total} =  \sum\limits_{i=1}^5(\mathcal{L}_{BCE}(\mathbf{S}_{i}, \mathbf{G})+\mathcal{L}_{SSIM}(\mathbf{S}_{i}, \mathbf{G}))
\end{array},
\end{equation}
where $\mathcal{L}_{total}$ denoted the hybrid loss, $\mathcal{L}_{BCE}$ and $\mathcal{L}_{SSIM}$ are BCE loss and SSIM loss, respectively. $\{\mathbf{S}_{i}\}_{i=1}^{5}$, and $\mathbf{G}$ represent the prediction map of each decoding stage and the ground truth.

\section{Experiments}\label{Experiments}
\subsection{Experimental Setup}\label{Experimental Setup}
\subsubsection{Dataset}
To give a comprehensive evaluation of our method, we conduct quantitative and qualitative experiments on a challenging strip steel surface defect detection dataset, namely SD-Saliency-900\cite{edrnet}. To be specific, the SD-Saliency-900 dataset consists of a training set and a testing set, each of which includes three distinct types of defects(\emph{i.e.,} inclusion, patches, and scratches). Each type of defect contains 300 images with a resolution of $200\times200$. The three types of defects are collected from real-world strip steel production processes, and their appearance is considerably different.
Besides, the SD-Saliency-900 dataset provides a pixel-wise annotation, where its training set contains 810 images and the testing dataset contains 900 images. In addition, to avoid overfitting, we augment the training set by horizontal flipping, which generates 1620 training images totally.

\subsubsection{Implementation Details}
Before the training process, we uniformly resize the training images to a spatial resolution of $368\times368$, and to enhance our MINet's robustness, we perform a random crop on the image, where we can obtain a $336\times336$ image. Here, we should note that the resolution of each input image in the testing process is $368\times368$. During the training phase, each image is normalized by subtracting the mean value 0.4669 and dividing by the variance value 0.2437. Besides, in terms of hardware environment, our MINet is trained and tested on a PC with an Intel(R) Core(TM) i9-9900X 3.50GHz CPU, 32 GB RAM, and an NVIDIA GTX 2080Ti GPU. To ensure a fair comparison, the computational efficiencies of other methods are all conducted within this hardware environment. In addition, during the training process, the parameters of our MINet are initialized by using the Xavier\cite{glorot2010understanding}. Meanwhile, we have employed Adam\cite{Adam} as our optimizer, and we set the initial learning rate and batch size to $4\times10^{-3}$ and 32, respectively. 
We train our MINet 92K iterations until convergency.

\subsubsection{Evaluation Criteria}
To evaluate the balance between accuracy and computational efficiency, we have chosen 8 key metrics to evaluate the accuracy of our MINet, including Mean Absolute Error (MAE), weighted F-measure (WF), overlapping ratio (OR), structure-measure (SM), Pratt's figure of merit (PFOM), Intersection over Union (IoU), Precision-Recall (PR) curve, and F-measure curve. Besides, to evaluate the computational efficiency of different methods, we adopt the following three metrics including the parameter size (Param), the floating-point operations (FLOPs), and the inference speed (Speed), which are measured in million (M), giga (G), and frames per second (FPS). Specifically, ``Param'' denotes the total parameters of the network, which is treated as an indicator of storage resource consumption. ``FLOPs'' refers to the floating-point operations, which is employed to quantify the computational complexity of the network. ``Speed'' represents the number of frames processed by the network per second, which can directly reflect the network?s inference speed.

\begin{figure*}[ht]
\centerline{\includegraphics[width=\textwidth]{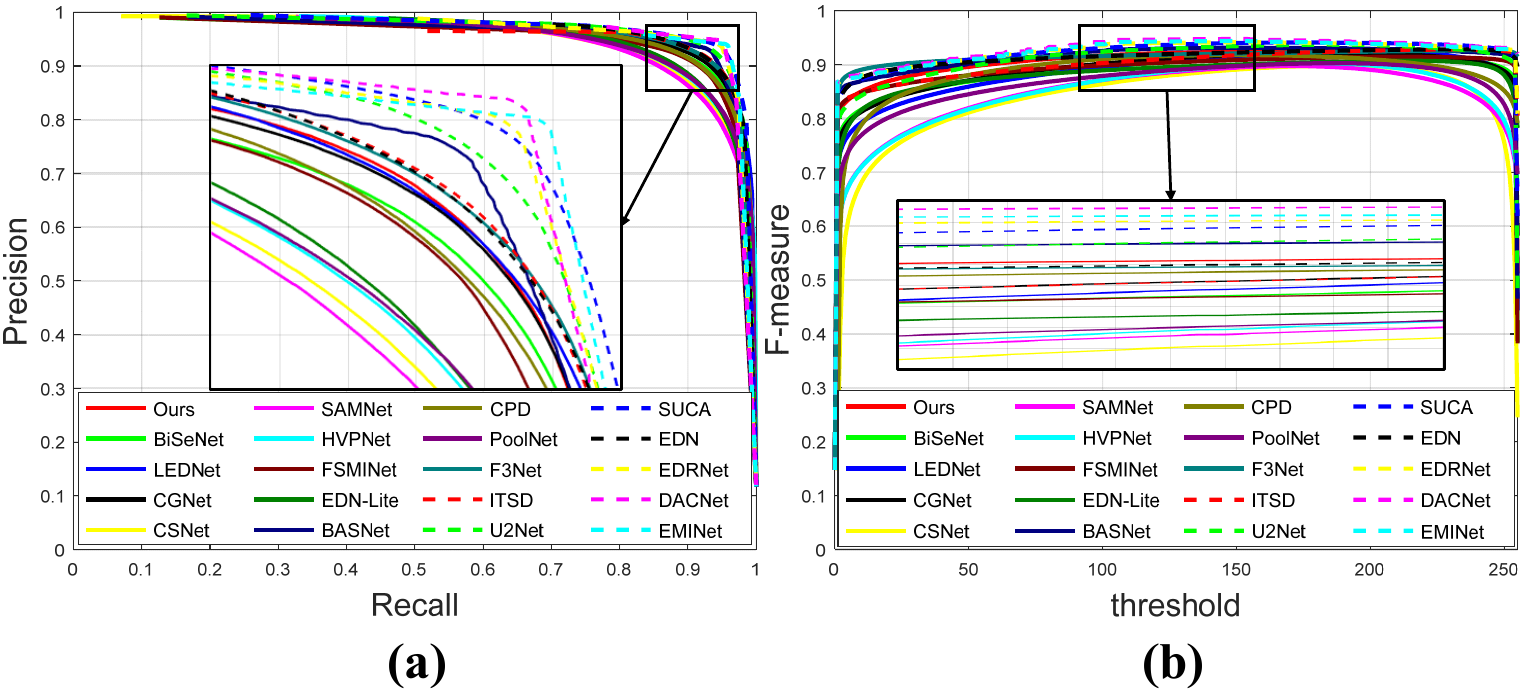}}
\caption{Quantitative evaluation of different networks. (a) PR curves, (b) F-measure curves on SD-Saliency-900 dataset.}
\label{PR-F-Curve}
\end{figure*}

\begin{table}[h!]
\centering
\caption{Quantitative comparisons with 19 state-of-the-art methods on SD-Saliency-900 dataset.}
\renewcommand{\arraystretch}{1.4}
\renewcommand{\tabcolsep}{0.6mm}
\scalebox{0.9}{
\begin{tabular}{c|ccc|cccccc}
\toprule
\multirow{2}[2]{*}{Methods} & Param & FLOPs & Speed & \multirow{2}[2]{*}{MAE$\downarrow$} & \multirow{2}[2]{*}{WF$\uparrow$} & \multirow{2}[2]{*}{OR$\uparrow$} & \multirow{2}[2]{*}{SM$\uparrow$} & \multirow{2}[2]{*}{PFOM$\uparrow$} & \multirow{2}[2]{*}{IoU$\uparrow$} \\
& (M)   & (G)   & (FPS) &       &       &       &       &       &  \\
\midrule
BASNet$_{19}$ & 87.06 & 127.40 & 57    & 0.0152 & 0.9092 & 0.8267 & 0.9276 & 0.8961 & 0.8450 \\
CPD$_{19}$   & 29.23 & 59.43 & 137   & 0.0211 & 0.8192 & 0.7749 & 0.9033 & 0.8856 & 0.7420 \\
PoolNet$_{19}$ & 52.51 & 117.00 & 70    & 0.0215 & 0.8476 & 0.7456 & 0.9022 & 0.8649 & 0.7554 \\
F3Net$_{20}$ & \textcolor[rgb]{ .267,  .447,  .769}{\textbf{25.54}} & \textcolor[rgb]{ .267,  .447,  .769}{\textbf{16.43}} & \textcolor[rgb]{ .439,  .678,  .278}{\textbf{217}} & 0.0150 & 0.9101 & 0.8401 & 0.9209 & 0.8832 & 0.8475 \\
ITSD$_{20}$  & 26.07 & \textcolor[rgb]{ .439,  .678,  .278}{\textbf{15.94}} & \textcolor[rgb]{ .267,  .447,  .769}{\textbf{208}} & 0.0153 & 0.9012 & 0.8099 & 0.9238 & 0.8967 & 0.8384 \\
U2Net$_{20}$ & 44.01 & 58.77 & 79    & 0.0143 & 0.9071 & 0.8134 & 0.9295 & 0.9088 & 0.8419 \\
SUCA$_{21}$  & 115.58 & 56.75 & 110   & \textcolor[rgb]{ .267,  .447,  .769}{\textbf{0.0125}} & 0.9191 & \textcolor[rgb]{ .267,  .447,  .769}{\textbf{0.8425}} & \textcolor[rgb]{ .267,  .447,  .769}{\textbf{0.9375}} & \textcolor[rgb]{ .267,  .447,  .769}{\textbf{0.9171}} & 0.8641 \\
EDN$_{22}$   & \textcolor[rgb]{ .439,  .678,  .278}{\textbf{21.83}} & 58.01 & 60    & 0.0149 & 0.9115 & 0.8308 & 0.9240 & 0.9021 & 0.8398 \\
\midrule
EDRNet$_{20}$ & 39.31 & 42.14 & 76    & 0.0130 & \textcolor[rgb]{ .267,  .447,  .769}{\textbf{0.9225}} & 0.8417 & \textcolor[rgb]{ .267,  .447,  .769}{\textbf{0.9375}} & 0.9133 & \textcolor[rgb]{ .267,  .447,  .769}{\textbf{0.8660}} \\
DACNet$_{21}$ & 98.39 & 142.71 & 54    & \textcolor[rgb]{ 1,  0,  0}{\textbf{0.0118}} & \textcolor[rgb]{ 1,  0,  0}{\textbf{0.9275}} & \textcolor[rgb]{ 1,  0,  0}{\textbf{0.8464}} & \textcolor[rgb]{ .439,  .678,  .278}{\textbf{0.9417}} & \textcolor[rgb]{ 1,  0,  0}{\textbf{0.9204}} & \textcolor[rgb]{ 1,  0,  0}{\textbf{0.8738}} \\
EMINet$_{21}$ & 99.13 & 139.71 & 52    & \textcolor[rgb]{ .439,  .678,  .278}{\textbf{0.0119}} & \textcolor[rgb]{ .439,  .678,  .278}{\textbf{0.9253}} & \textcolor[rgb]{ .439,  .678,  .278}{\textbf{0.8447}} & \textcolor[rgb]{ 1,  0,  0}{\textbf{0.9422}} & \textcolor[rgb]{ .439,  .678,  .278}{\textbf{0.9173}} & \textcolor[rgb]{ .439,  .678,  .278}{\textbf{0.8736}} \\
\midrule
Ours  & \textcolor[rgb]{ 1,  0,  0}{\textbf{0.28}} & \textcolor[rgb]{ 1,  0,  0}{\textbf{0.30}} & \textcolor[rgb]{ 1,  0,  0}{\textbf{721}} & 0.0169 & 0.8945 & 0.8115 & 0.9207 & 0.8943 & 0.8123 \\
\midrule
\midrule
BiSeNet$_{18}$ & 13.24 & 7.76  & \textcolor[rgb]{ .439,  .678,  .278}{\textbf{709}} & \textcolor[rgb]{ .267,  .447,  .769}{\textbf{0.0178}} & 0.8803 & 0.7848 & 0.9155 & 0.8868 & 0.8075 \\
LEDNet$_{19}$ & 0.92  & 2.95  & 408   & 0.0182 & 0.8782 & 0.7730 & \textcolor[rgb]{ .439,  .678,  .278}{\textbf{0.9196}} & \textcolor[rgb]{ .439,  .678,  .278}{\textbf{0.8929}} & 0.8035 \\
CGNet$_{20}$ & \textcolor[rgb]{ .267,  .447,  .769}{\textbf{0.49}} & 1.78  & 605   & \textcolor[rgb]{ .439,  .678,  .278}{\textbf{0.0172}} & \textcolor[rgb]{ .267,  .447,  .769}{\textbf{0.8825}} & \textcolor[rgb]{ .267,  .447,  .769}{\textbf{0.7854}} & \textcolor[rgb]{ .267,  .447,  .769}{\textbf{0.9191}} & \textcolor[rgb]{ .267,  .447,  .769}{\textbf{0.8906}} & \textcolor[rgb]{ .267,  .447,  .769}{\textbf{0.8099}} \\
\midrule
CSNet$_{20}$ & \textcolor[rgb]{ 1,  0,  0}{\textbf{0.14}} & \textcolor[rgb]{ .267,  .447,  .769}{\textbf{0.72}} & 495   & 0.0309 & 0.7659 & 0.7026 & 0.8839 & 0.8443 & 0.6804 \\
SAMNet$_{21}$ & 1.33  & \textcolor[rgb]{ .439,  .678,  .278}{\textbf{0.54}} & \textcolor[rgb]{ .267,  .447,  .769}{\textbf{621}} & 0.0278 & 0.8161 & 0.7082 & 0.8885 & 0.8276 & 0.7116 \\
HVPNet$_{21}$ & 1.23  & 1.12  & 567   & 0.0271 & 0.8231 & 0.7110 & 0.8938 & 0.8442 & 0.7247 \\
FSMINet$_{22}$ & 3.56  & 11.79 & 62    & \textcolor[rgb]{ .267,  .447,  .769}{\textbf{0.0178}} & \textcolor[rgb]{ .439,  .678,  .278}{\textbf{0.8899}} & \textcolor[rgb]{ .439,  .678,  .278}{\textbf{0.8008}} & 0.9142 & 0.8803 & \textcolor[rgb]{ 1,  0,  0}{\textbf{0.8142}} \\
EDN-Lite$_{22}$ & 1.80  & 1.04  & 517   & 0.0204 & 0.8674 & 0.7741 & 0.8988 & 0.8497 & 0.7779 \\
\midrule
Ours  & \textcolor[rgb]{ .439,  .678,  .278}{\textbf{0.28}} & \textcolor[rgb]{ 1,  0,  0}{\textbf{0.30}} & \textcolor[rgb]{ 1,  0,  0}{\textbf{721}} & \textcolor[rgb]{ 1,  0,  0}{\textbf{0.0169}} & \textcolor[rgb]{ 1,  0,  0}{\textbf{0.8945}} & \textcolor[rgb]{ 1,  0,  0}{\textbf{0.8115}} & \textcolor[rgb]{ 1,  0,  0}{\textbf{0.9207}} & \textcolor[rgb]{ 1,  0,  0}{\textbf{0.8943}} & \textcolor[rgb]{ .439,  .678,  .278}{\textbf{0.8123}} \\
\bottomrule
\end{tabular}%
}
\begin{tablenotes}
\footnotesize
\item The best three results in each column have been highlighted in \textcolor[rgb]{ 1,  0,  0}{red}, \textcolor[rgb]{ .439,  .678,  .278}{green}, and \textcolor[rgb]{ .267,  .447,  .769}{blue}, respectively.
\item $\downarrow$ means the smaller, the better, while $\uparrow$ denotes the larger, the better.
\end{tablenotes}
\label{Qun1}%
\end{table}%

\subsection{Comparison with state-of-the-art}\label{Comparison with state-of-the-art}
\begin{figure*}[h]
\centerline{\includegraphics[width=\textwidth]{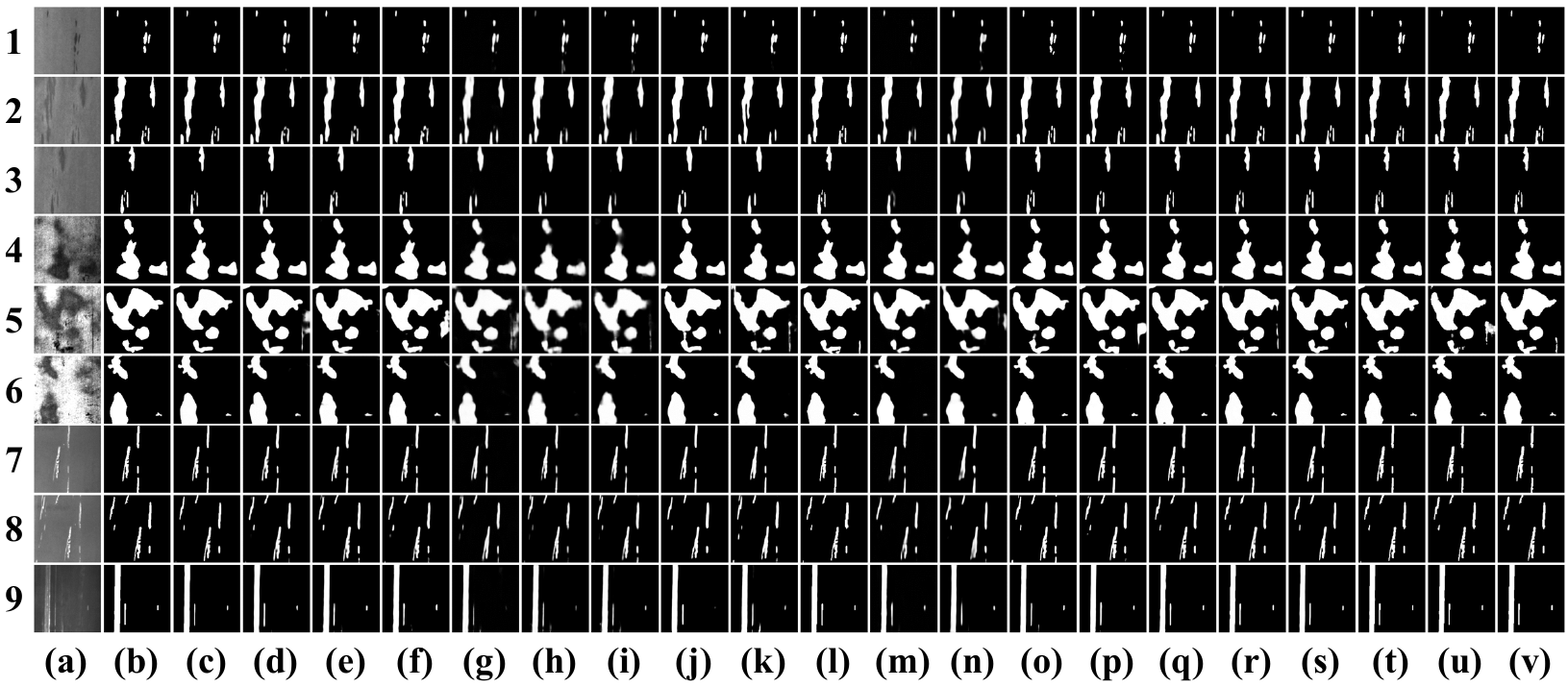}}
\caption{Qualitative comparisons of different models. (a) Image, (b) GT, (c) Ours, (d)BiSeNet\cite{BiSeNet}, (e)LEDNet\cite{LEDNet}, (f)CGNet\cite{CGNet}, (g) CSNet\cite{CSNet}, (h) SAMNet\cite{SAMNet}, (i) HVPNet\cite{HVPNet}, (j) FSMINet\cite{FSMINet}, (k) EDN-Lite\cite{EDN}, (l) BASNet\cite{BASNet}, (m) CPD\cite{CPD}, (n) PoolNet\cite{PoolNet}, (o) F3Net\cite{F3Net}, (p) ITSD\cite{ITSD}, (q) U2Net\cite{U2Net}, (r) SUCA\cite{SUCA}, (s) EDN\cite{EDN}, (t) EDRNet\cite{edrnet}, (u) DACNet\cite{dacnet}, (v) EMINet\cite{eminet}.}
\label{Visual_Com_all}
\end{figure*}

In this section, in Table~\ref{Qun1}, Fig.~\ref{PR-F-Curve}, and Fig.~\ref{Visual_Com_all}, we make a comparison between our method and 19 state-of-the-art methods, where 8 traditional methods are designed for natural scenes (BASNet\cite{BASNet}, CPD\cite{CPD}, PoolNet\cite{PoolNet}, F3Net\cite{F3Net}, ITSD\cite{ITSD}, U2Net\cite{U2Net}, SUCA\cite{SUCA}, and EDN\cite{EDN}), 3 methods major in strip steel surface defect detection (EDRNet\cite{edrnet}, DACNet\cite{dacnet}, and EMINet\cite{eminet}), 3 real-time segmentation methods (BiSeNet\cite{BiSeNet}, LEDNet\cite{LEDNet}, and CGNet\cite{CGNet}), 5 lightweight detection methods (CSNet\cite{CSNet}, SAMNet\cite{SAMNet}, HVPNet\cite{HVPNet}, FSMINet\cite{FSMINet}, and EDN-Lite\cite{EDN}).





\subsubsection{Network comparison}
We conduct the comparison between our MINet and state-of-the-art methods, as shown in Table~\ref{Qun1}. The PR curves and F-measure curves are shown in Fig.~\ref{PR-F-Curve}. From Table~\ref{Qun1}, we can find that our MINet exhibits the highest processing speed (721 FPS) and the lowest computational cost (0.28M parameters and 0.30G FLOPs) when compared with traditional methods. For instance, the MAE value of our MINet is 0.0169, and the MAE of the top-level DACNet\cite{dacnet} is 0.0118. 
However, the parameters of our MINet are only 0.28\% of DACNet's yet the processing speed of MINet is 13.4 times that of DACNet. In real-world scene, the spatial resolution of images is usually larger than 1024$\times$1024. Therefore, we also test the inference speed of DACNet and our MINet on the 1024$\times$1024 images, where the inference speed of the two methods is 3FPS and 95FPS, respectively. According to this result, we can see that our model can better meet real-time processing requirements when compared with DACNet. 
Meanwhile, from Fig.~\ref{PR-F-Curve} and Table~\ref{Qun1}, we can see that the proposed MINet achieves competitive detection accuracy when compared with traditional detection methods.

Besides, compared with the real-time or lightweight methods, our MINet exhibits the best performance in terms of PR curves, F-measure curves, MAE, weighted F-measure (WF), overlapping ratio (OR), structure-measure (SM), and Pratt's figure of merit (PFOM). Meanwhile, in terms of computational efficiency including Param, FLOPs, and Speed, our MINet boasts the second-lowest number of parameters and the fastest inference speed. Specifically, as shown in Table~\ref{Qun1}, our MINet achieves an SM value of 0.9207, while CGNet attains an SM value of 0.9191. Besides, our MINet achieves a Speed of 721FPS, which is faster than that of CGNet. The Params and FLOPs of the proposed MINet are 0.28M and 0.30G, which are both lower than that of CGNet. Furthermore, when compared with the fastest real-time model BiSeNet (709 FPS), the Speed of our MINet is still faster. Similarly, when compared with CSNet, which has the lowest parameters (0.14 M), the IoU of our MINet is 0.1319 higher, but we are only with an additional 0.14M parameters. 
According to the aforementioned experiments, we can say that our MINet has successfully achieved a well balance between detection accuracy and computational efficiency.

Furthermore, the visual comparison results are presented in Fig.~\ref{Visual_Com_all}. We can see that our method shown in Fig.~\ref{Visual_Com_all}(c) achieves a comparable performance when compared with the traditional methods targeting natural scene images and surface defect detection such as EDRNet, SUCA, \emph{etc}. When compared with the lightweight or real-time methods such as EDN-Lite and FSMINet, as shown in Fig.~\ref{Visual_Com_all}(k, j), our method shown in Fig.~\ref{Visual_Com_all}(c) presents better performance. More specifically, in view of suppressing cluttered backgrounds (\emph{e.g.}, 5$^{th}$ example), our MINet can well suppress background regions and present accurate detection results. Besides, for the 7$^{th}$ example, our MINet shown in Fig.~\ref{Visual_Com_all}(c) exhibits more accurate boundary details when compared to other lightweight or real-time methods, such as BiSeNet, LEDNet, and CSNet presented in Fig.~\ref{Visual_Com_all}(d, e, g).


\subsection{Ablation analysis}\label{Ablation analysis}
In this section, to validate the key components of our method including MI-based real-time backbone, the architecture of MI module, the feature combination types, and the loss function, we conduct extensive ablation studies, as shown in  Table~\ref{Qun2} and Table~\ref{QunAB}.

\begin{table}[h!]
\centering
\caption{Quantitative comparisons with different backbones.}
\footnotesize
\renewcommand{\arraystretch}{1.4}
\renewcommand{\tabcolsep}{0.6mm}
\scalebox{0.9}{
\begin{tabular}{c|ccc|cccccc}
\toprule
\multirow{2}[2]{*}{Methods} & Param & FLOPs & Speed & \multirow{2}[2]{*}{MAE$\downarrow$} & \multirow{2}[2]{*}{WF$\uparrow$} & \multirow{2}[2]{*}{OR$\uparrow$} & \multirow{2}[2]{*}{SM$\uparrow$} & \multirow{2}[2]{*}{PFOM$\uparrow$} & \multirow{2}[2]{*}{IoU$\uparrow$} \\
& (M)   & (G)   & (FPS) &       &       &       &       &       &  \\
\midrule
Ours  & 0.28  & 0.30  & 721   & 0.0169 & 0.8945 & 0.8115 & 0.9207 & 0.8943 & 0.8123 \\
\midrule
VGG16 & 15.81 & 45.85 & 93    & 0.0135 & 0.9119 & 0.8241 & 0.9352 & 0.9048 & 0.6522 \\
ResNet50 & 31.96 & 14.96 & 186   & 0.0153 & 0.8998 & 0.8055 & 0.9303 & 0.9045 & 0.7993 \\
\midrule
SqeeezeNet & 1.27  & 1.79  & 586   & 0.0183 & 0.8767 & 0.7837 & 0.9146 & 0.8806 & 0.8135 \\
MobileNet & 7.47  & 2.51  & 443   & 0.0174 & 0.8887 & 0.7900 & 0.9214 & 0.8915 & 0.8023 \\
EfficientNet & 0.94  & 0.24  & 451   & 0.0178 & 0.8863 & 0.7964 & 0.9182 & 0.8891 & 0.8043 \\
\bottomrule
\end{tabular}%
}
\begin{tablenotes}
\footnotesize
\item $\downarrow$ means the smaller, the better, while $\uparrow$ denotes the larger, the better.
\end{tablenotes}
\label{Qun2}%
\end{table}%

\subsubsection{Backbone comparison}\label{Backbone comparison}
To validate the effect of different backbones, we design five variants of our MINet, as shown in Table~\ref{Qun2}. Specifically, we replace our MI-based real-time backbone with two large-scale backbones (ResNet50\cite{resnet} and VGG16\cite{vgg}) and three real-time backbones (SqueezeNet\cite{squeezenet}, MobileNet\cite{mobilenet}, and EfficientNet\cite{efficientnet}), which are defined as ``ResNet50'', ``VGG16'', ``SqueezeNet'', ``MobileNet'', and ``EfficientNet'', respectively.


Formally, when compared with a large-scale backbone such as VGG16, the Param and FLOPs of our method are only 1.8\% and 0.7\% of VGG16's. However, the SM value of our method is only 0.0145 lower than VGG16's. The case of ResNet50 is similar to that of VGG16. From the observations, we can prove the effectiveness and high efficiency of our MI-based real-time backbone when compared with the large-scale backbone. Besides, when compared with one of the top-level real-time backbones SqueezeNet, in terms of MAE, WF, OR, SM, and PFOM, our method performs best. Meanwhile, in terms of computational efficiency metrics including Param, FLOPs, and Speed, the Param, FLOPs, and Speed of our method are 22.0\%, 16.8\%, and 1.23 times of SqueezeNet's, respectively. From the analysis, we can firmly prove the effectiveness and high efficiency of our MI-based real-time backbone.


\begin{table}[h!]
\centering
\caption{Quantitative comparisons with different variants.}
\footnotesize
\renewcommand{\arraystretch}{1.4}
\renewcommand{\tabcolsep}{0.6mm}
\scalebox{0.9}{
\begin{tabular}{c|ccc|cccccc}
\toprule
\multirow{2}[2]{*}{Methods} & Param & FLOPs & Speed & \multirow{2}[2]{*}{MAE$\downarrow$} & \multirow{2}[2]{*}{WF$\uparrow$} & \multirow{2}[2]{*}{OR$\uparrow$} & \multirow{2}[2]{*}{SM$\uparrow$} & \multirow{2}[2]{*}{PFOM$\uparrow$} & \multirow{2}[2]{*}{IoU$\uparrow$} \\
& (M)   & (G)   & (FPS) &       &       &       &       &       &  \\
\midrule
Ours  & 0.28  & 0.30  & 721   & 0.0169 & 0.8945 & 0.8115 & 0.9207 & 0.8943 & 0.8123 \\
\midrule
w/o Multi & 0.28  & 0.30  & 721   & 0.0182 & 0.8847 & 0.7912 & 0.9177 & 0.8889 & 0.7937 \\
S-PW  & 0.27  & 0.29  & 941   & 0.0194 & 0.8752 & 0.7689 & 0.9120 & 0.8717 & 0.8021 \\
w/o CI & 0.15  & 0.20  & 770   & 0.0188 & 0.8797 & 0.7819 & 0.9152 & 0.8794 & 0.8099 \\
Concat & 1.15  & 0.98  & 677   & 0.0165 & 0.8975 & 0.8240 & 0.9208 & 0.8960 & 0.8153 \\
Sum   & 0.65  & 0.59  & 758   & 0.0182 & 0.8786 & 0.7767 & 0.9137 & 0.8728 & 0.8041 \\
\midrule
w focal & 0.28  & 0.30  & 721   & 0.0195 & 0.8797 & 0.7870 & 0.9137 & 0.8835 & 0.8025 \\
w Lov{\'a}sz & 0.28  & 0.30  & 721   & 0.0177 & 0.8877 & 0.7904 & 0.9179 & 0.8877 & 0.8058 \\
w importance & 0.28  & 0.30  & 721   & 0.0193 & 0.8723 & 0.7527 & 0.9133 & 0.8762 & 0.7967 \\
w Abhishek & 0.28  & 0.30  & 721   & 0.0295 & 0.7921 & 0.7562 & 0.8489 & 0.7501 & 0.8571 \\
w/o SSIM & 0.28  & 0.30  & 721   & 0.0180 & 0.8837 & 0.7839 & 0.9203 & 0.8899 & 0.8299 \\
\bottomrule
\end{tabular}%
}
\begin{tablenotes}
\footnotesize
\item $\downarrow$ means the smaller, the better, while $\uparrow$ denotes the larger, the better.
\end{tablenotes}
\label{QunAB}%
\end{table}%

\subsubsection{MI module's architecture}
To validate the effectiveness of the architecture of the MI module, we design three variants, as shown in Table~\ref{QunAB}. Concretely, according to Fig.~\ref{MI module}, to demonstrate the effectiveness of the multi-scale strategy of the ``Extraction'' stage, we first utilize four DWConvs with the same dilation rate to replace the original four DWConvs with different dilation rates, which is marked as ``w/o Multi''. Secondly, we replace the concatenation operation in the ``Fusion'' stage with the summation operation, which is denoted as ``S-PW''. Thirdly, according to Eq.~\ref{eq4}, we remove the PWConv, which is replaced by a 1$\times$1 convolutional layer (channel number = c). This variant is written as ``w/o CI''.  From the results presented in Table~\ref{QunAB}, we can find that our method performs best in terms of detection accuracy, which clearly demonstrates the effectiveness and rationality of the design of our method. Besides, we can also find an interesting thing, where ``w/o Multi'' is with the same computational efficiency as our method. This can further prove the rationality of the deployment of the multi-scale strategy in our method.


\subsubsection{Feature combination types}
In Sec.~\ref{Preliminary}, we discuss three main types of methods which are used to fuse the multi-scale features. Considering that $3^{rd}$ method has many configurations, we only conduct experiments on the first two methods, which are marked as ``Concat'' and ``Sum'', respectively. From Table~\ref{QunAB}, we can find that the ``Concat'' yields better performance than our method. However, it also incurs a significant computational cost, where the Param and FLOPs of ``Concat'' are $4.1$ times and $3.3$ times of ours, respectively. For the variant ``Sum'', it performs worse than ``Concat'' and our method. This also proves that the summation operation can result in the dilution of features. From the above experiments, we can demonstrate the effectiveness and rationality of the design of our MI module.


\subsubsection{Loss function}
To validate the effectiveness of our loss function, we design a variant ``w/o SSIM'', where we remove the SSIM loss from our hybrid loss. Furthermore, we replace the loss function of our MINet with four existing loss functions, including focal loss\cite{focal}, Lov{\'a}sz-Softmax loss\cite{lovasz}, importance-aware loss\cite{importance}, and Abhishek loss\cite{Adhishek}, respectively. The four variants are denoted as ``w focal'', ``w Lov{\'a}sz'', ``w importance'', and ``w Abhishek'', respectively. From the results illustrated in Table~\ref{QunAB}, we can find that ``w/o SSIM'' performs worse than our method. This indicates that SSIM loss can effectively improve the detection accuracy without extra computational cost, which is friendly to real-time detection methods. Moreover, from Table~\ref{QunAB}, we can find that our model performs better than the existing four variants including ``w focal'', ``w Lov{\'a}sz'', ``w importance'', and ``w Abhishek''. This clearly demonstrates the effectiveness of our loss function.

\subsection{Practical implications and challenges}
\subsubsection{Practical implications}
In industrial production processes, various defects are inevitable. Therefore, automatic defect detection has been introduced \cite{ni2021attention,dong2019pga}. Here, the primary application of our MINet is strip steel surface defect detection. Moreover, as a preprocessing method, SOD-based defect detection\cite{edrnet, dacnet} has also been successfully employed in various industrial tasks, such as defect classification\cite{hu2016surface} and defect inspection\cite{xu2019defect}. Therefore, our MINet can also be applied to the aforementioned industrial tasks. Besides, the primary innovation of this work involves the exploration of a novel lightweight multi-scale feature interaction method. Embarking on this, we constructed our multi-scale interactive (MI) module. Our MI module structure is with a single input and a single output, which can be treated as a fundamental module and plugged into existing methods. For instance, by mimicking the ResNet50 architecture, we build our MI-based real-time backbone. As shown in Table III, with only a marginal sacrifice in detection accuracy, our MINet has a significant boost in inference speed when compared with the ResNet50-based network. This implies that our MI-based real-time backbone can be applied in numerous vision tasks, such as object detection \cite{song2023steel}, and semantic segmentation \cite{dong2019pga}.


\subsubsection{Challenges}
In practical industrial scenarios, the working environment of RGB cameras is highly variable, where images often exhibit various issues such as uneven illumination, overexposure, and low light conditions. Besides, we notice that multiple similar objects in the images also pose big challenges. Recently, some researchers \cite{song2022novel} attempted to introduce visible cameras, depth cameras, and thermal infrared cameras to tackle the aforementioned challenges. Here, thermal infrared cameras can mitigate issues arising from variations in light, while depth cameras can help address problems related to background interference. Therefore, in our future work, the combination of depth and thermal infrared data will be a promising solution to more industry applications.

\section{Conclusion}\label{Conclusion}
In this paper, we first present a novel real-time module to acquire multi-scale features, which is called multi-scale interactive (MI) module. The key point of MI module lies in that we can organize the multi-scale feature extraction and fusion into a unified framework with few parameters, small FLOPs, and fast inference speed. 
Embarking on the MI module, we build our real-time backbone, which consists of five encoder blocks and can be used to extract multi-scale deep features. To deal with the task of surface defect detection, we propose the encoder-decoder architecture based network, namely MINet. 
Extensive experimental results on the SD-Saliency-900 dataset demonstrate that our MINet has much smaller computational cost, better real-time performance, and competitive detection accuracy when compared with the cutting-edge methods. Furthermore, compared with existing large-scale backbones and real-time backbones, our MI-based real-time backbone has presented a well balance between computational efficiency and detection accuracy. In our future work, to adapt to a broader range of industrial scenarios, we will attempt to extend our MINet to multi-modal data, such as depth and thermal images. Besides, we plan to utilize some other lightweight techniques, such as pruning and re-parameterization, to build more advanced lightweight detection models.

\bibliographystyle{IEEEtran}
\bibliography{reference}

\begin{IEEEbiography}
[{\includegraphics[width=1in,height=1.25in,clip,keepaspectratio]{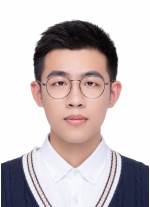}}]
{Kunye Shen} received the B.E. degree from Hangzhou Dianzi University, Hangzhou, China, in 2023. He is currently pursuing the Ph.D. degree in the School of Communication and Information Engineering, Shanghai University, Shanghai, China. His research interests include computer vision, and visual saliency analysis.
\end{IEEEbiography}
\begin{IEEEbiography}
[{\includegraphics[width=1in,height=1.25in,clip,keepaspectratio]{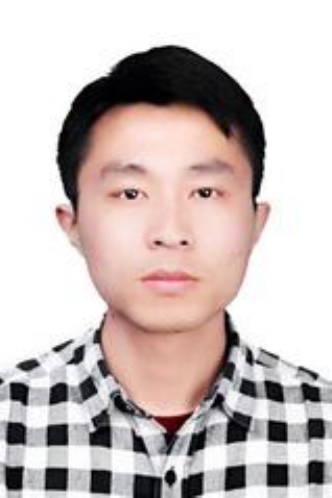}}]
{Xiaofei Zhou} received the Ph.D. degree from Shanghai University, Shanghai, China, in 2018. He is currently an Associate Professor with the School of Automation, Hangzhou Dianzi University, Hangzhou, China. His research interests include saliency detection, video segmentation, image enhancement, and defect detection.
\end{IEEEbiography}
\begin{IEEEbiography}
[{\includegraphics[width=1in,height=1.25in,clip,keepaspectratio]{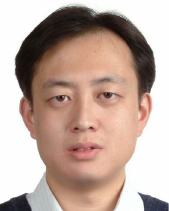}}]
{Zhi Liu} (Senior Member, IEEE) received the B.E. and M.E. degrees from Tianjin University, Tianjin, China, and the Ph.D. degree from the Institute of Image Processing and Pattern Recognition, Shanghai Jiao Tong University, Shanghai, China, in 1999, 2002 and 2005, respectively. He is currently a Professor at the School of Communication and Information Engineering, Shanghai University, Shanghai, China. From August 2012 to August 2014, he was a Visiting Researcher with the SIROCCO Team, IRISA/INRIA-Rennes, France, with the support by EU FP7 Marie Curie Actions. He has published more than 200 refereed technical papers in international journals and conferences. His research interests include image/video processing, machine learning, computer vision, and multimedia communication. He is an Area Editor of \textit{Signal Processing: Image Communication} and served as a Guest Editor for the special issue on \textit{Recent Advances in Saliency Models, Applications and Evaluations} in \textit{Signal Processing: Image Communication}.
\end{IEEEbiography}
\end{document}